\documentclass[12pt]{spieman}  
\usepackage{amsmath,amsfonts,amssymb}
\usepackage{graphicx}
\usepackage{setspace}
\usepackage{tocloft}
\usepackage{lineno}

\usepackage{xcolor} 
\newcommand{\highlight}[1]{#1}
\newcommand{\blue}[1]{#1}

\makeatletter
\renewcommand{\figurename}{\textbf{Fig}}
\renewcommand{\fnum@figure}{\figurename~\textbf{\thefigure}}
\makeatother

\title{\textbf{{Image class translation: visual inspection of class-specific hypotheticals and classification based on translation distance}}}

\author[a]{Mikyla K. Bowen}
\author[b, c, *]{Jesse W. Wilson}
\affil[a]{College of Natural Sciences, Colorado State University, Colorado, United States of America}
\affil[b]{School of Biomedical and Chemical Engineering, Colorado State University, Colorado, United States of America}
\affil[c]{Department of Electrical and Computer Engineering, Colorado State University, Colorado, United States of America}

\cftpagenumbersoff{figure}
\cftpagenumbersoff{table} 
\begin{document} 
\maketitle

\begin{abstract}
\section*{Purpose:} A major barrier to the implementation of artificial intelligence for medical applications is \blue{automated CNNs'} lack of explainability and high confidence \blue{for} incorrect decisions, specifically with out-of-domain samples. We propose a generalization of image translation networks for image classification and demonstrate \blue{translation networks'} potential as a more interpretable alternative to conventional black-box classifiers. 
\section*{Approach:} We train an \blue{image-to-image} network to translate an input image to class-specific hypotheticals, and then compare these with the input, both visually and quantitatively. Translation distances, the degree of alteration needed to conform to one class or another, are examined for clusters and trends, and used as \blue{a} simple low-dimensional feature vector for classification. 
\section*{Results:} On melanoma/benign dermoscopy images, a translation distance classifier achieved 80\% accuracy using only a 2-dimensional feature space (versus 85\% for a conventional CNN using a $\sim 62,000$-dimensional feature space). Visual inspection of rendered images \blue{revealed dataset biases, like more scalebars in melanoma photographs than in benign lesions}. Image distributions in translation distance space revealed a natural separation along the lines of dermatologist decision to biopsy, rather than between malignant and benign. On bone marrow cytology images, translation distance classifiers outperformed a conventional CNN in both 3-class (92\% accuracy vs 89\% for CNN) and 6-class (90\% vs 86\% for CNN) scenarios.
\section*{Conclusions:} This proof-of-concept shows the potential for image-to-image translation to go beyond artistic/stylistic changes and to expose dataset biases, perform dimension reduction and dataset visualization, and in some cases, potentially outperform conventional end-to-end CNN classifiers.
\end{abstract}


{\noindent \footnotesize\textbf{*}Jesse W. Wilson,  \linkable{jesse.wilson@colostate.edu} }

\begin{spacing}{2}   

\section{Introduction}
\label{sect:intro}  
Recent advances in machine learning such as deep convolutional neural networks (CNNs) have brought about revolutionary new capabilities for automated object recognition and classification~\cite{NIPS2012_c399862d}. While the accuracies yielded are impressive, mistakes and misclassifications are still an area of concern, especially in medical applications. CNNs trained on images from one clinic perform poorly on images from another clinic, known as the out-of-domain problem, and trivial changes to an image, e.g., brightness, lighting, or rotation can fool state-of-the-art classifiers, even when these changes are incorporated as augmentations to the training data~\cite{Alcorn19}. \blue{Compounding that}, misclassifications with high confidence scores can mislead experts into making the wrong call~\cite{Tschandl20}. For these reasons, clinicians have been reluctant to make use of AI in routine practice.

One possible solution is to make a better CNN that does not suffer from the out-of-domain problem, a difficult task that has not yet been solved~\cite{Young21}. Another solution lies in interpretable/explainable AI (xAI). The idea is if a human can examine a network's decision-making process they can decide whether to trust the network’s output for a specific case and use that information to make a more informed decision{~\cite{PlougThomas2020Tfdo, xAI22, Chanda2024}}. Most of these approaches work with a conventional CNN classifier and analyze hidden layer activations~\cite{Strobelt18}. However, even with advances like class activation maps~\cite{7780688}, which {highlight pixels} that contributed to {a} decision, a CNN still cannot explain what {high-level characteristics were involved.}

\begin{figure}
\centering \includegraphics[scale=.5] {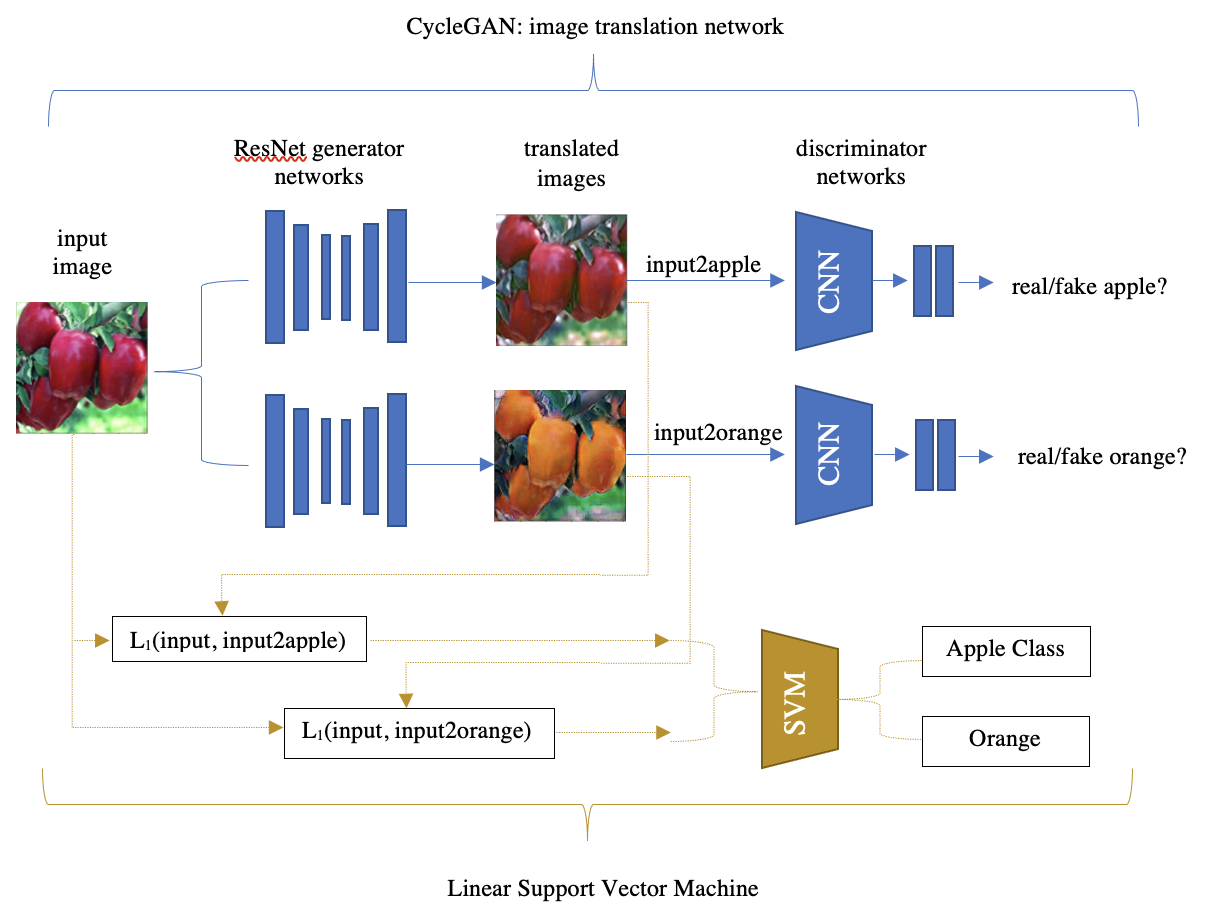} 
\caption{Image2image classification network architecture. An image is fed through an image translation network such as a CycleGAN above or StarGAN, where the generator and discriminator compete to make realistic images of each class. The resulting images are used for visual inspection, to compute translation ($L_1$) distances, which can used for classification or used in a translation distance classifier.} 
\label{fig:architecture} 
\end{figure}

Here, we propose to use generative-adversarial~\cite{Goodfellow14} image2image~\cite{Zhu17,choi2018stargan} networks to translate input images into $K$ class-specific hypotheticals, where $K$ is the number of classes\blue{. We then compare the hypotheticals with the input for qualitative visual changes and quantitative translation distance (Fig.~\ref{fig:architecture}). Qualitative visual changes are observed manually and can include both biologically-meaningful changes such as changes in morphology, color, or texture, or artifacts such as absence or presence of a scalebar or vignetting (darkening at the edges of the image). Quantitative translation distance calculations are described below.}. Conventional applications of image2image are modality transform (I2I-MT), e.g. style transfer for artistic purposes~\cite{chen2018cartoonGAN} {or} translating between medical imaging modalities~\cite{lei2019mri2ct}. The novelty here is using image2image for \emph{class} transform (I2I-CT) and analyzing, both visually and quantitatively, the changes needed to fit a given image into the $K$ possible target classes.

Unlike a conventional black-box classifier, visual inspection of I2I-CT hypotheticals can draw an observer’s attention to case-specific defining characteristics and dataset trends that can bias conventional classifiers, such as biologically-irrelevant artifacts like scalebars and vignetting, which significantly degrade generalizability of CNN classifiers trained to identify melanoma~\cite{Winkler2021, Sies2021}. Quantitative comparison through $L_1$ distance between the input and class-specific hypotheticals reduces each image to a length-$K$ vector, $d$, of translation distances. Each dimension of this simple feature vector represents the amount of alteration needed to fit into a class, and can be used to visualize dataset distributions (through, e.g. scatterplots in translation distance space) or as inputs to a simple downstream classifier (e.g. a linear classifier, fully-connected neural network, or support vector machine (SVM)~\cite{Pedregosa}).

Our approach is related to previous works in generative counterfactuals~\cite{DeGrave}, abnormal-to-normal translation (ANT-GAN)~\cite{8950113}, and visual attribution ANT-GAN (VANT-GAN)~\cite{ZIA2022112}. While saliency and activation mapping approaches (e.g. Grad-CAM~\cite{Selvaraju_2017_ICCV}) are limited to highlighting which pixels were most important, generative image2image approaches render higher-level factors influencing a classifier's decision and depict case-specific defining features of disease. For example, in the counterfactuals approach, an extension of Explanation by Progressive Exaggeration~\cite{Singla2020Explanation}, an image2image network learns to produce realistic variations (counterfactual images) that alter a classifier's decision. A visual inspection of such counterfactuals then reveals whether the classifier is sensitive to the same type of features as human domain experts. In the ANT-GAN approach, an image2image CycleGAN learns to translate images having potential anomalies (e.g. tumors) to hypothetical healthy counterparts. Subtracting these two or, in the case of VANT-GAN, directly synthesizing the difference image reveals abnormal regions. More recent work in this space has been to incorporate advances in stable diffusion image generation and large language models~\cite{Siddiqui2025}. The key differences between counterfactuals and ANT-GAN can be summed up as follows. In counterfactuals, a classifier is central, and image2image is used to analyze the classifier; in ANT-GAN, image2image is central. In counterfactuals, images are produced that either increase or decrease a classifier's estimated likelihood of membership in a certain class; in ANT-GAN, an image is translated to a possible target class, where ideally translation to the true class leaves the image unaltered.

In this manuscript, we describe image2image class transform (I2I-CT), an extension and generalization of ANT-GAN, where (1) target classes are not limited to just an abnormal versus normal distinction; (2) generated images are not subtracted from source images, rather, visual inspection is performed in like manner to counterfactuals to support interpretation of classifier results and uncover training dataset biases; and (3) translation distances are quantified and evaluated as interpretable low-dimensional representations for images, used for visualizing distributions of images, or as a simple feature vectors for classification. 

We first demonstrate I2I-CT on two-class $(K=2)$ problems using CycleGAN: a trivial apples versus oranges task, and then the difficult melanoma versus benign pigmented lesion task. Though a translation distance classifier in these $K=2$ examples does not outperform a conventional CNN (EfficientNet), we include these results to establish proof of principle and highlight the value of manual inspection of class-hypothetical images and translation distance distributions. Finally, we move into $K=3$ and $K=6$ problems with a StarGAN and demonstrate significant overall improvement in accuracy versus a conventional CNN, although we emphasize that classification accuracy improvement over state-of-the-art end-to-end CNNs is not the primary focus of this work.

\section{Methods}
\blue{Image2Image Class trasform (I2I-CT)} uses an image translation network $G_i$, (e.g. CycleGAN, StarGAN, etc.), to translate an input image $x$ to produce a set of $K$ corresponding class-specific hypothetical images $y_i=G_i(x)$, where $i\in\{1,2,...,K\}$. These images are then compared visually with the original input image, and quantitatively by image translation distance.

\subsection{Image Translation Networks}

Image2image networks learn the mappings between classes~\cite{Isola16} using a generator  discriminator. For example, $G_i$ learns to produce an image in a target domain $i$, conditioned on an input image $x$. \blue{Conversely} $D_i$ provides adversarial feedback on whether the result is a realistic sample from the target domain. The adversarial loss ensures a matching distribution for the translated images and the target class\cite{Zhu17}. Additional loss terms often found in image2image include identity loss $\lambda_I$ and cycle consistency loss $\lambda_{\rm cycle}$. Identity loss $\lambda_I$, penalizes in-class alterations, i.e. images translated into their own domain should not change. Cycle consistency loss $\lambda_{\rm cycle}$, encourages reversible transformations. These parameters influence the image generation into the same class as well as the limits or lack thereof for generation into opposing classes. Image2image approaches vary depending on whether they rely on supervised training with paired datasets (pix2pix\cite{Isola16}), unsupervised training (CycleGAN\cite{Zhu17}, NCE/CUT\cite{park2020contrastivelearningunpairedimagetoimage}), support $K>2$ domains (StarGAN\cite{STARGAN}), or produce multiple target domain variants (e.g. MUNIT\cite{Huang_2018_ECCV}, DRIT\cite{DRIT}).

Our experiments used a CycleGAN for $K=2$ experiments, and StarGAN for $K>3$, as these approaches do no require paired input images, and instead rely on overall domain specific characteristics.

\highlight{\subsection{Image Translation Distance }}

\highlight{
For proof-of-concept, we simply used pixel-wise mean average difference $d_i = \| x - G_i(x)\|_1$, \blue{also referred to as $L_1$ norm or $L_1$ distance}. However, other metrics such as Structural Similarity Index Metric~\cite{Brunet2012_SSIM} or convolutional encoder perceptual distances~\cite{Zhang_2018_CVPR} could in principle be used. Together, the translation distances for each class-specific hypothetical reduce an image to a simple $K$-dimensional feature vector $d=\{d_1, d_2, ..., d_K\}$ that can be used for viewing distributions of images (e.g. through scatterplots) in a dataset or directly for classification. The translation distance feature vector has a straightforward interpretation, requiring no technical expertise: each dimension is proportional to the amount of alteration needed to fit an image into a particular class. In other words, images should have small translation distances (ideally, zero) for classes to which they belong and larger translation distances for classes to which they do not belong.
}

{\subsubsection{CycleGAN network, K=2}}
The CycleGAN model is a TensorFlow implementation~\cite{Keras20} of the original PyTorch CycleGAN~\cite{Zhu17} for 256x256 images which also added dropout to the residual layers. Both the generators and discriminators used the optimizer Adam with a learning rate of $\alpha=2\times 10^{-4}$ and $\beta_1=0.5$. Models were trained for 60,600 iterations.

Initially, we used the default values of $\lambda_I=5$ and $\lambda_{\rm cycle=10}$. We tuned the two parameters as follows. Translation distance vectors $d$ were fed into a SVM, and results were assessed with receiver operating characteristic (ROC) area under the ROC (AUROC) as a measurement of the CycleGAN's ability to separate images by class. We then maximized AUROC by tuning $\lambda_I$ and $\lambda_{\rm cycle}$.

\subsubsection{StarGAN network, K\textgreater{}2}
For $K>2$ scenarios, we used the original version of StarGAN~\cite{STARGAN}, and added identity loss. As with the CycleGAN, we tuned $\lambda_I$ by evaluating the models based on the resulting accuracy and the confusion matrices of an SVM on the translation distance vectors as well as visual image quality. The Bone Cytology experiment used a $\lambda_I=.001$. (Note: this is $10^4\times$ smaller than in the CycleGAN because of the difference between the two codebases in using \textit{mean} absolute error versus \textit{sum} absolute error in their loss functions). The model trained for a minimum of 300,000 iterations for the experiments. Increasing the number of classes and samples during training and testing required more iterations to achieve comparable image quality. We used a batch size of 16, an image size of 128, and the remaining default parameters stayed the same as the original StarGAN.

\subsection{Simple Classifiers }
The translation distance is a reduction of the original input image into a $K$-dimensional feature vector, where the smallest value of the $K$-class options is the predicted class. To account for cases where there is minimal differences between classes or where the translation distance might not necessarily correlate with the correct class, we additionally trained a simple classifier, specifically a support vector machine (SVM). This is trained using the translation distances from the image generation training set, and then tested on the remaining images. \blue{We initially used a simple linear SVM, but we later also compared additional classifiers such as SVMs with different kernels, logistic regression, translation distance, and a multilayer perceptron.}

{\subsection{Baseline CNN classifier}}
For each dataset, a baseline CNN, EfficientNetB3~\cite{Tan19}, pre-trained on ImageNet~\cite{deng2009imagenet} was trained on experiment data using transfer learning. Training augmentations included vertical and horizontal flip, random brightness and contrast, optical distortion, grid distortion, elastic transform, contrast limited adaptive histogram equalization, cutout, and color jitter, using the Albumentation package~\cite{Buslaev_2020}. 
To adjust for class imbalance, we applied class weights with the datasets with a traditional 80/20 split. This was calculated as $w_i = \left( \frac{1}{c_i} \right) \cdot \left( \frac{\sum_{j} c_j}{2} \right)$. For the datasets with already an equal number of images for each class during, the CNN used the same dataset for comparison purposes with the StarGAN. 
The baseline CNNs trained using an early stopping callback to monitor the validation set loss with a patience of 200 epochs.

\section{Experiments }
We conducted four class transformation (I2I-CT) experiments: 1) 2-class CycleGAN on apples/oranges, 2) 2-class CycleGAN pigmented {lesions, 3) 3-class StarGAN on bone marrow cytology, and 4) 6-class StarGAN on bone marrow cytology. For each, the performance of the {image2image} approach was compared with the baseline CNN end-to-end classifier. For the 2-class problems, to measure performance while accounting for class imbalance we used the area under the receiver operator curve (AUROC)~\cite{HanleyROC}. For the multi-class problems, we looked at traditional overall accuracy and then individual class accuracy in a confusion matrix. }

\subsection{Experiment 1: Apples and Oranges, K=2} 
To demonstrate the concept of image translation and classification using translation distance, we first applied I2I-CT to a simple task: identifying apples and oranges. This was a non-domain specific task that highlights clear categorical differences between classes.

\subsubsection{Data}

The apples/oranges dataset is a subset of the larger ImageNet dataset~\cite{ImageNet} compiled by the CycleGAN authors~\cite{Zhu17}, from which we randomly selected a training set of 995 apple and 1019 orange images; our test set had 266 apple and 248 orange images.

\subsection{Experiment 2: Pigmented Lesions, K=2}
Classifying melanoma versus benign pigmented lesions is relatively more difficult than apples versus oranges, but is a more clinically relevant task. Applying I2I-CT, allows visual inspection of class specific features and more transparency and intepretability for mistakes and classification where classes are not as easily distinguishable.

\subsubsection{Data} 
The pigmented lesions dataset consists of dermoscopic photographs from the International Skin Imaging Collaboration (ISIC) 2020 challenge dataset~\cite{ISIC20}, which contains images collected from the Hospital Clínic de Barcelona, the Medical University of Vienna, Memorial Sloan Kettering Cancer Center, Melanoma Institute Australia, the University of Queensland, and the University of Athens Medical School. There are 30,928 benign images and 584 malignant images. To mitigate the class imbalance problem~\cite{class16}, we used a random subset of 2,584 images with an 80/20 train/test split (453 malignant and 1614 benign images in the training set and 131 malignant and 386 benign images in the test set). The baseline CNN and CycleGAN networks 
used the same data splits for performance comparisons.

While training used the ISIC data from 2020, out-of-distribution performance was tested with malignant images from the 2016 and 2017 ISIC challenges~\cite{ISIC16, ISIC17}, and additional benign images from the ISIC 2020 challenge that were excluded from the training subset~\cite{ISIC20}.

 \subsubsection{Experiment Adjustments }
Given the difficult task of distinguishing more similar classes, we performed hyperparameter tuning to maximize the separation of classes. The SVM used a weight of three to adjust for the class imbalance, and performance improvements were based on the linear SVM.

 \subsection{Experiment 3: Bone Marrow Cytology, K=3 }
The third experiment expanded testing to $K=3$, using images from the Bone Marrow Smear Cytology dataset. This experiment focused on classification of cells with domain specific characteristics, specifically those with different morphological features. Extending to $K=3$, also shifts the problem from a binary to a multiclass classification task, which lays the groundwork for future practical applications with an increased number of domains. 

 \subsubsection{Data }

 For the multi-class experiments, we used the Bone Marrow Cytology in Hematologic Malignancies Data~\cite{bone_data}, which represents 21 different cell classifications. For the $K=3$ model, we selected three categories: blasts, erythroblasts, and lymphocytes. These classes were selected because of the large availability of images. The number of training images was set at 5,000 randomly-selected images for each class (15,000 total training images), to eliminate the class imbalance problem. {The remaining 50,610 images of each class} were used for testing (6,973 blasts, 22,395 erythroblasts, and 21,242 lymphocytes). 

 \subsubsection{Experiment Adjustments }
The CycleGAN architecture only supports unpaired translation between two classes. For $K>2$, we used a StarGAN, which utilizes a conditional GAN, allowing us to specify $K=3$ as the number of classes. Images were randomly selected and then fed into the StarGAN using the file directory based on classes.

 \subsection{Experiment 4: Bone Marrow Cytology, K=6 }

Experiment 4 built on the $K=3$ experiment, adding in three additional classes. The purpose of this experiment was to demonstrate classification ability when $K>3$, moving towards larger more applicable classification tasks. This allowed us to consider compounding classification difficulties for certain classes that scale with the task as well as overall impacts on global accuracy. 

\subsubsection{Data }

For the $K=6$ model, we added segmented neutrophils, plasma cells, and promyelocytes. These were selected because they also met the requirements of having more than 5,000 images each; 5,000 images from each of these classes were added to the images of the previous three classes, and the remaining images added to the test set {(24,434 segmented neutrophils, 2,629 plasma cells, and 11,994 promylelocytes). }

 \subsubsection{Experiment Adjustments }

 An additional experiment leveraged the entire $K=6$ dataset with an 80/20 train test split using a weighted sampler to mitigate the class imbalance problem. We used PyTorch's WeightedRandomSampler during the data processing phase to calculate and create class weights. The inverse frequency of each class was given as the weight for each image based on its target. $w_i = \frac1{c_i}$, where $c_i$ is the class frequency. 

Furthermore, we evaluated auxiliary classifiers' performances against a traditional classifier, EfficientNet, as a baseline. We compared SVMs with linear, RBF, and poly kernels, logistic regression, smallest translation distance, and a multilayer perceptron.

\section{Results}
\subsection{Experiment 1: Apples and Oranges, K=2 }

The CycleGAN+SVM using translation distance vectors achieved an AUROC of .9926 and .9545 on the training and test sets, respectively. In comparison, the end-to-end CNN classified apples and oranges with an AUROC of .9922 and .9825 on the training and test sets. However by changing to $\lambda_{\rm cycle}=0$, CycleGAN+SVM AUROC increased to .9971 and .9705 on the training and test sets. The traditional CNN only outperforms the CycleGAN+SVM approach by a marginal 1.2\%, as measured by the test set AUROC. 

\begin{figure*}
\begin{center}
\includegraphics[width=1\linewidth]{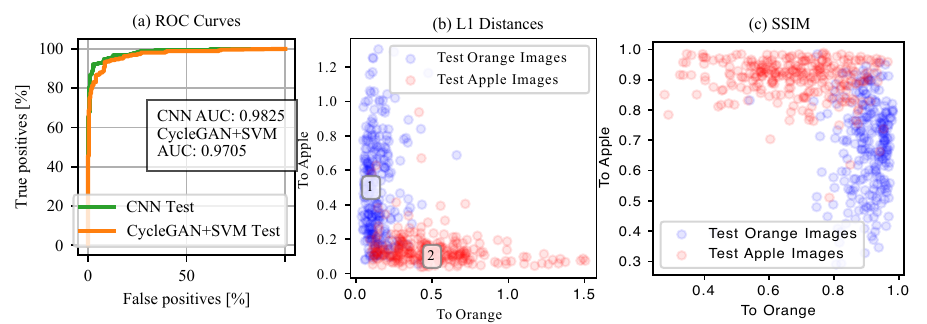}
\end{center}
   \caption{(a) ROC Curve of an {end-to-end} CNN {(green line)} and the {translation distance SVM (orange line)}. (b) Scatterplot of {test set translation distances. \blue{Oranges cluster in Region 1, apples cluster in Region 2.}(c) Scatterplot of structural similarity index measure (SSIM).} }
\label{fig:a20_plots}
\end{figure*}

\begin{figure*}
\begin{center}
\includegraphics[width=1\linewidth]{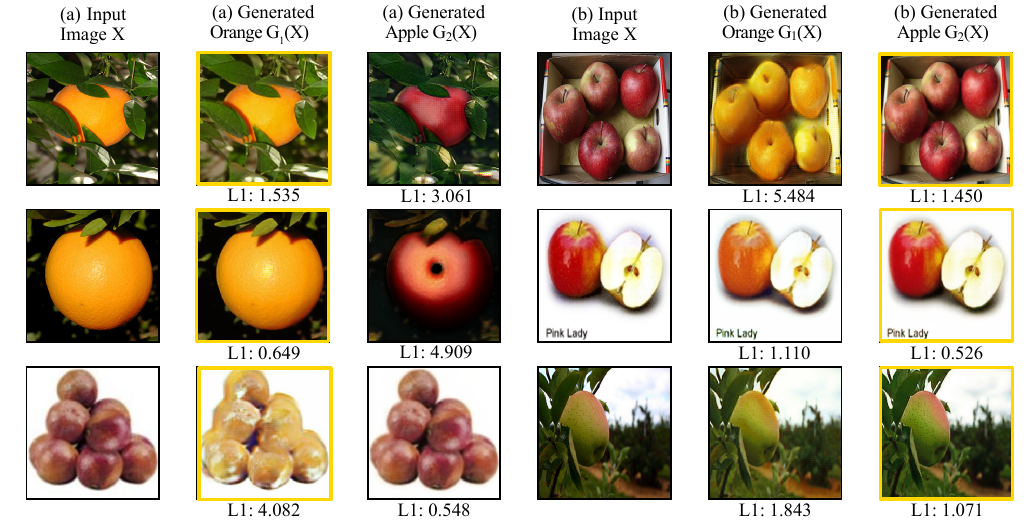}
\end{center}
   \caption{Images highlighted correspond to the correct image domain.(a) Input orange images translated into the orange domain, {$G_1(X)$}, and apple domain, {$G_2(X)$}. (b) Input apple images translated into the orange domain, {$G_1(X)$}, and apple domain, {$G_2(X)$}.}
\label{fig:apple_oranges}
\end{figure*}

The separability of apples and oranges in translation distance space can be seen in the \blue{scatterplots (Fig.~\ref{fig:a20_plots}(b, c))}.
 {Figure~\ref{fig:apple_oranges} shows a selection of images produced from the apples ({class 1}) to oranges ({class 2}) CyleGAN model.
The top row of Fig.~\ref{fig:apple_oranges} shows images of reasonable translation quality that were correctly classified by {a translation distance SVM}. The middle row shows images that failed to translate morphology in the opposing class but {a translation distance SVM still obtained the} correct classification. The bottom row shows uncharacteristic source images that led to translation or classification failures. Visual inspection shows features such as fruit color, background color, and texture that were changed or added to the original image to produce hypothetical images for each class. }

\subsection{Experiment 2: Pigmented Lesions, K=2 }

The translation distance SVM achieved a test set AUROC of 0.8044 as compared to the end-to-end CNN which achieved 0.8536. Both are slightly higher than dermatologist visual inspection, where AUROC ranges from 0.671~\cite{BRINKER201947} to 0.79~\cite{Haenssle}. State-of-the-art end-to-end CNNs report test set AUROCs ranging from 0.654 to 0.949, though performance on external datasets is often dramatically lower~\cite{Young21}.

\begin{figure*}
\begin{center}
\includegraphics[width=1\linewidth]{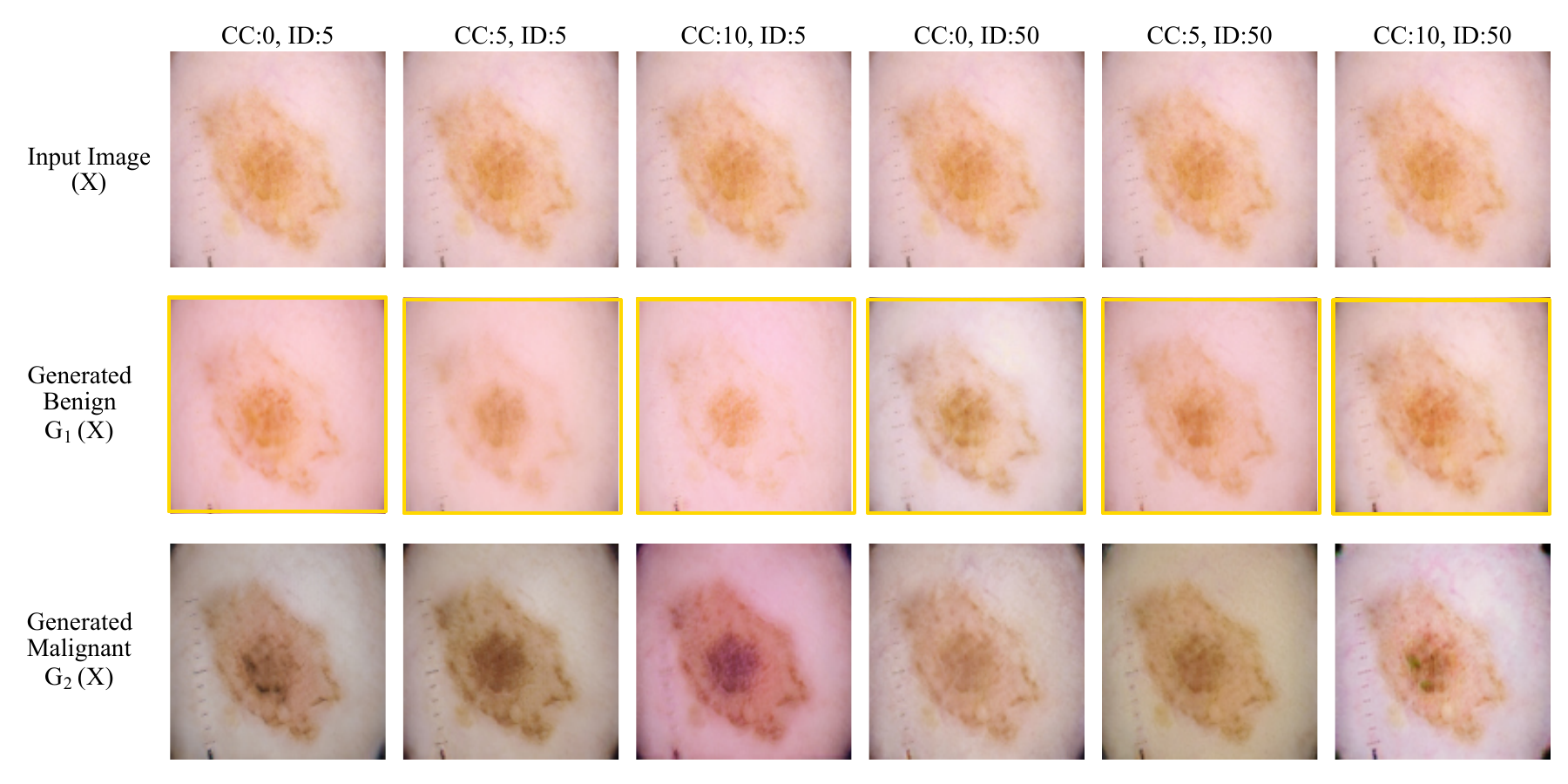}
\end{center}
   \caption{Image panel of input benign images (X) translated into the benign class, $G_1$, and the malignant class,$G_2$ using different lambda values for cycle consistency (CC) and identity (ID). Yellow highlight indicates in-class translation.}
\label{fig:benign_samples_ID}
\end{figure*}

\begin{figure*}
\begin{center}
\includegraphics[width=1\linewidth]{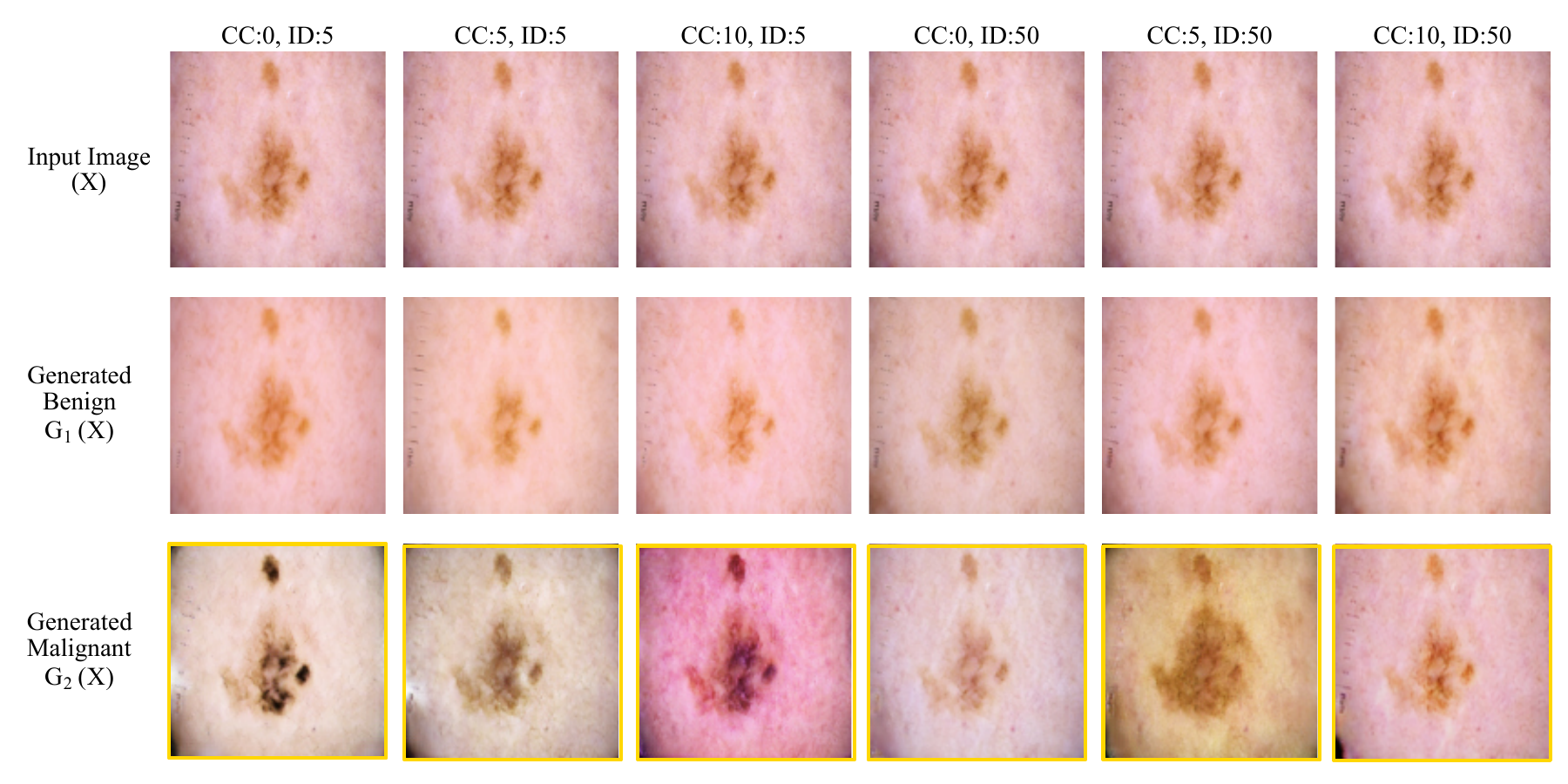}
\end{center}
   \caption{Image panel of input malignant images (X) translated into the benign class, $G_1$, and the malignant class, $G_2$ using different lambda values for cycle consistency (CC) and identity (ID). Yellow highlight indicates in-class translation.}
\label{fig:malignant_samples_ID}
\end{figure*}

\begin{figure*}
\begin{center}
\includegraphics[width=1\linewidth]{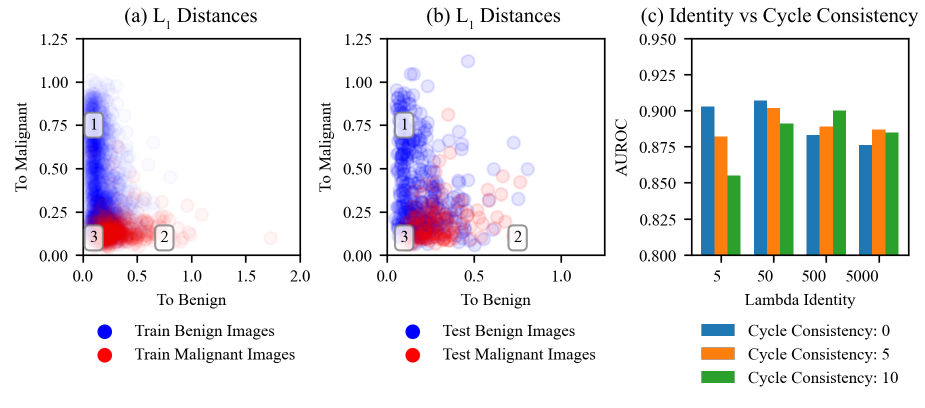}
\end{center}
   \caption{ (a) Scatterplot of training images $L_1$ distance coordinates. (b) Scatterplot of test images $L_1$ distance coordinates. Numbered regions correspond to (1) benign, (2) malignant, and (3) ambiguous based on similar translation distance. (c) AUROC Metrics for Hyperparameter tuning cycle consistency and identity loss.}
\label{fig:m2b_plots}
\end{figure*}

The CycleGAN+SVM results described in this section use $\lambda_I=5$ and $\lambda_{\rm cycle=0}$, which were found to be the best parameters after evaluating models with different combinations of $\lambda_I$ and $\lambda_{\rm cycle}$. We note that identity loss is necessary for preventing the network from changing an image if it is already in the correct domain. Image panels generated with different $\lambda_I$ and $\lambda_{\rm cycle}$ are shown in Figs.~\ref{fig:benign_samples_ID} and~\ref{fig:malignant_samples_ID}. Additionally, nonzero cycle consistency loss turned out to prevent any meaningful alterations to the images. AUROC for different $\lambda_I$ and $\lambda_{\rm cycle}$ is shown in Fig.~\ref{fig:m2b_plots}(c).

\begin{figure}
\begin{center}
\includegraphics[scale=1]{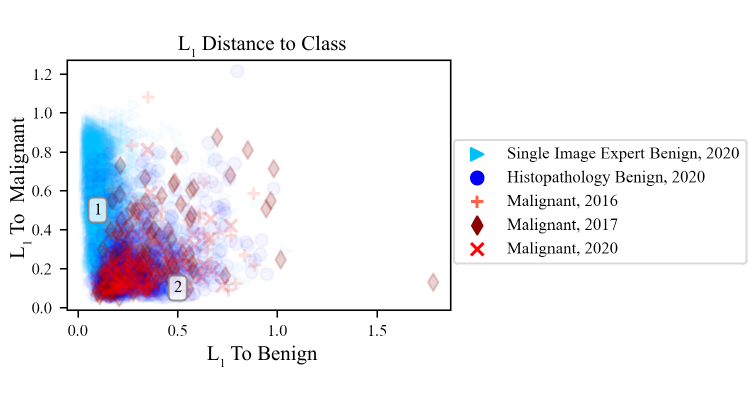}
\end{center}
   \caption{$L_1$ distances pairs on an expanded {out-of-training-distribution dataset, separated by class, year, and diagnosis confirmation type. Benign images are separated by method used for diagnosis: single image expert consensus or histopathology review, and all malignant images are determined through histopathology. Images in region 1 are expected to be benign, and those in region 2 are expected to be malignant, based on translation distance. Single image expert consensus images group in region 1 as expected, however images requiring histopathology, regardless of final diagnosis type fall in region 2.}}
\label{fig:histopathology}
\end{figure}

The translation distance scatterplots (a) and (b) in Fig.~\ref{fig:m2b_plots} show that the separation between the malignant and benign pigmented lesions is not as pronounced as in apples versus oranges (Fig.~\ref{fig:a20_plots}). Three regions in translation distance space can be identified: (1) contains benign lesions that are well-separated from the majority of melanomas; (2) contains melanomas that are well-separated from the majority of benign lesions; and (3) contains overlap between benign and malignant lesions. Upon further examination, testing with a larger number of benign images and malignant images from multiple years of the ISIC database (Fig.~\ref{fig:histopathology}), we found that region (1) consists mostly of images labeled ``Single Image Expert Benign" as opposed to ``Histopathology Benign" -- in other words, these images were considered unambiguously benign by a dermatologist, not requiring a biopsy to confirm. Furthermore, we find that region (1) tends to exclude melanomas, even with a test set expanded to include images from other years of the ISIC database.

\begin{figure*}
\begin{center}
\includegraphics[width=1\linewidth]{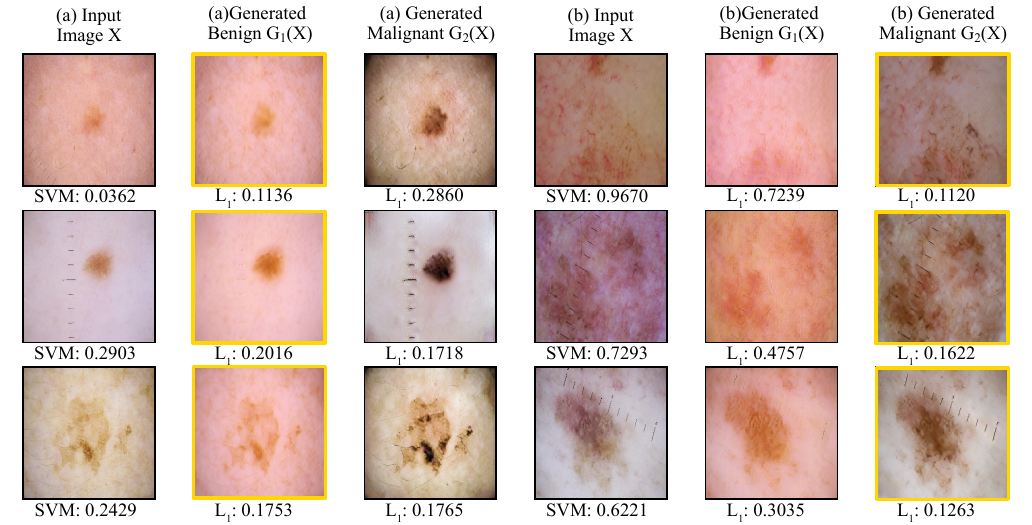}
\end{center}
   \caption{(a) Input benign and (b) input malignant images correctly classified by the CycleGAN+SVM at a .5 threshold. Yellow highlight indicates in-class translation.}
\label{fig:SVM_images}
\end{figure*}

For visual inspection, we first show a selection of images (Fig.~\ref{fig:SVM_images}) that were correctly classified by the {translation distance SVM} at a decision threshold of 0.5. In these cases, rendered images from the correct class tend to look more like the source image in terms of the pigmented lesion's color (brown/black), texture, and morphology. However, the generators also tend to alter other characteristics such as background skin color (in these cases, pale versus pink), scalebar markings, and vignetting (darkening around the edges due to the imaging optics or lighting conditions). From this, it would appear that the generators have learned to prefer rendering benign lesions with a pink background skin tone, and melanomas with a pale skin tone, scalebar markings, and vignetting. This suggests the presence of these artifacts may be biased towards one class or another in the training data, and may be a source of error in training conventional classifiers.

\begin{figure}
\centering
\includegraphics[width=1\linewidth] {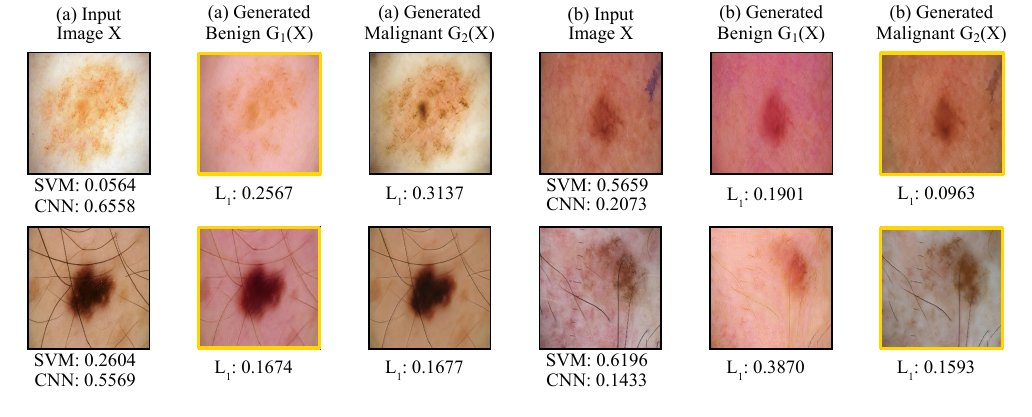}
   \caption{(a) Input benign images and (b) input malignant images incorrectly classified by a traditional CNN at .5, but correctly classified by {feeding translations distances into an SVM to predict classes using a decision boundary} (CycleGAN+SVM) at .5 threshold. CNN values closer to 1 and 0 indicate high confidence of being malignant and benign, respectively. Values in the middle are less confident predictions for either class. Yellow highlight indicates in-class translation.}
\label{fig:CNN_images}
\end{figure}

Secondly, a selection of images misclassified by the {baseline} CNN is shown in Fig.~\ref{fig:CNN_images}. In these cases, though the {translation distance SVM} does not always arrive at the correct classification, the generated images provide insight into artifacts and possible causes for misclassification. For example, the benign images in Fig.~\ref{fig:CNN_images} show the same tendency of the network to consider benign images as having a more pink background tone and melanomas as having  lighter skin tone and vignetting. As discussed above, this may reflect biases in the training dataset. For the malignant images in  Fig.~\ref{fig:CNN_images} (b), while the traditional CNN gave a low probability of melanoma, the generated malignant images appear visually much closer to the source images input malignant lesions, as is also reflected in the translation distances. {In these cases, a CNN may lead to a missed cancer diagnosis, but visual inspection of I2I-CT hypotheticals would raise concern.}

\begin{figure}
\centering
\includegraphics[width=1\linewidth] {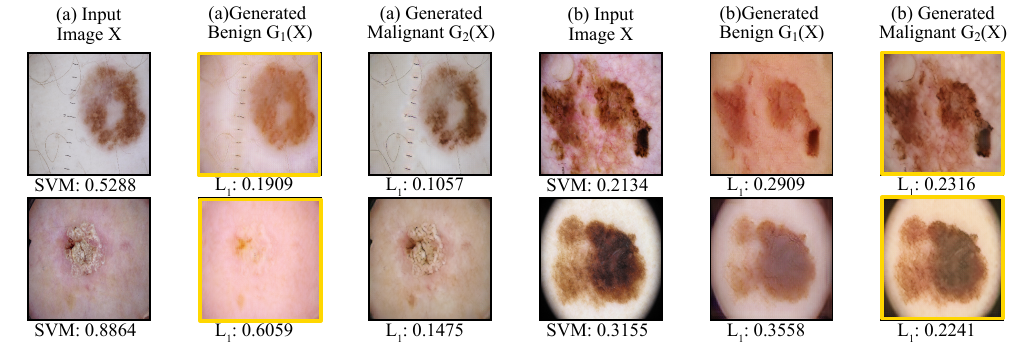}
   \caption{(a) Input benign images and (b) input malignant images incorrectly classified by the CycleGAN+SVM at a .5 threshold. Yellow highlight indicates in-class translation.}
\label{fig:SVM_missed_images}
\end{figure}

\begin{figure}
\centering
\includegraphics[width=1\linewidth] {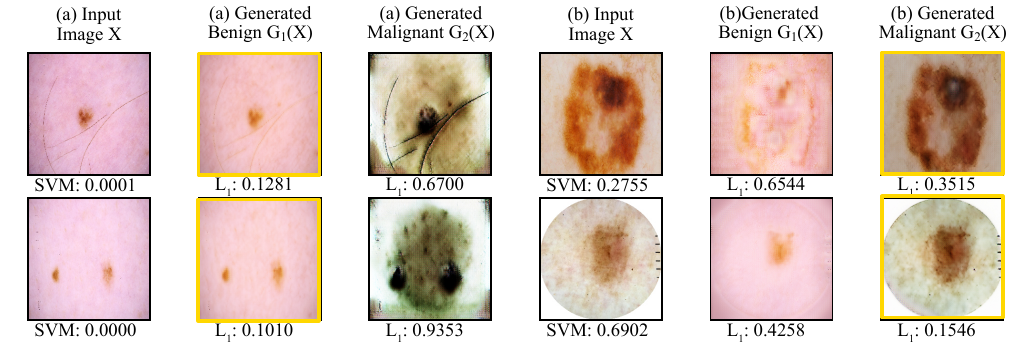}
   \caption{(a) Hallucinated artifacts in $G_B(X)$ with input benign images. (b) Hallucinated artifacts in $G_A(X)$ with input malignant images. Yellow highlight indicates in-class translation.}
\label{fig:Failure_images}
\end{figure}

Thirdly, a selection of images misclassified by the {translation distance SVM} is shown in Fig.~\ref{fig:SVM_missed_images}. In these cases, a visual inspection affords more room for interpretation and making an informed decision than would a black-box classifier. Consistent with earlier observations, it appears that benign lesions (Fig.~\ref{fig:SVM_missed_images} (a)) with pale skin background, scalebar markings, and vignetting tend to be misclassified as melanomas. In the case of malignant lesions (Fig.~\ref{fig:SVM_missed_images} (b)), though the SVM incorrectly labeled these benign, the source images are more visually similar to the malignant renderings.

 Finally, a selection of generator failures/hallucinations are shown in 
Fig.~\ref{fig:Failure_images}, in which rendered images in one domain or another show severe deviations from the source image. However, in each of these cases, both the translation distance SVM and visual inspection lead to correct classification. Note again the tendency towards pink skin tone in the rendered benign images and pale skin and vignetting in the rendered malignant images.

 \subsection{Experiment 3: Bone Marrow Cytology, K=3 }
 Utilizing the {$K=3$ subset of Bone Marrow Cytology in Hematologic Malignancies, we achieved 95\% training accuracy, and a 92\% test accuracy, predicting classes using the smallest translation distance. This was a marginal improvement over the baseline CNN, which achieved an 89\% test accuracy. Confusion matrices of both networks are shown in Fig.~\ref{fig:Confus_CNN}, showing more pronounced improvements, especially in the blast category, which had a 10 percentage point accuracy improvement over the EfficientNet.  Although blasts and lymphocytes continue to be difficult to resolve, I2I-CT translation distance cut the number of these misclassifications by roughly a factor of two compared to the baseline CNN.}

Compared to other networks' performances, ResNexT50 and EfficientNetV2, on this dataset our StarGAN+SVM achieved the same or improved accuracy for all three classes in the HematoNet results\cite{HematoNet22}. Similarly, our baseline EfficientNet was on par for all three classes. 
\begin{figure}
\begin{center}
\includegraphics[width=1\linewidth] {Confusion_matrices.pdf}
\end{center}
   \caption{{Confusion matrices for (a) EfficientNet and (b) smallest translation distance on $K=3$ cytology subset.}}
\label{fig:Confus_CNN}
\end{figure}

\begin{figure}
\begin{center}
\includegraphics[scale=0.7] {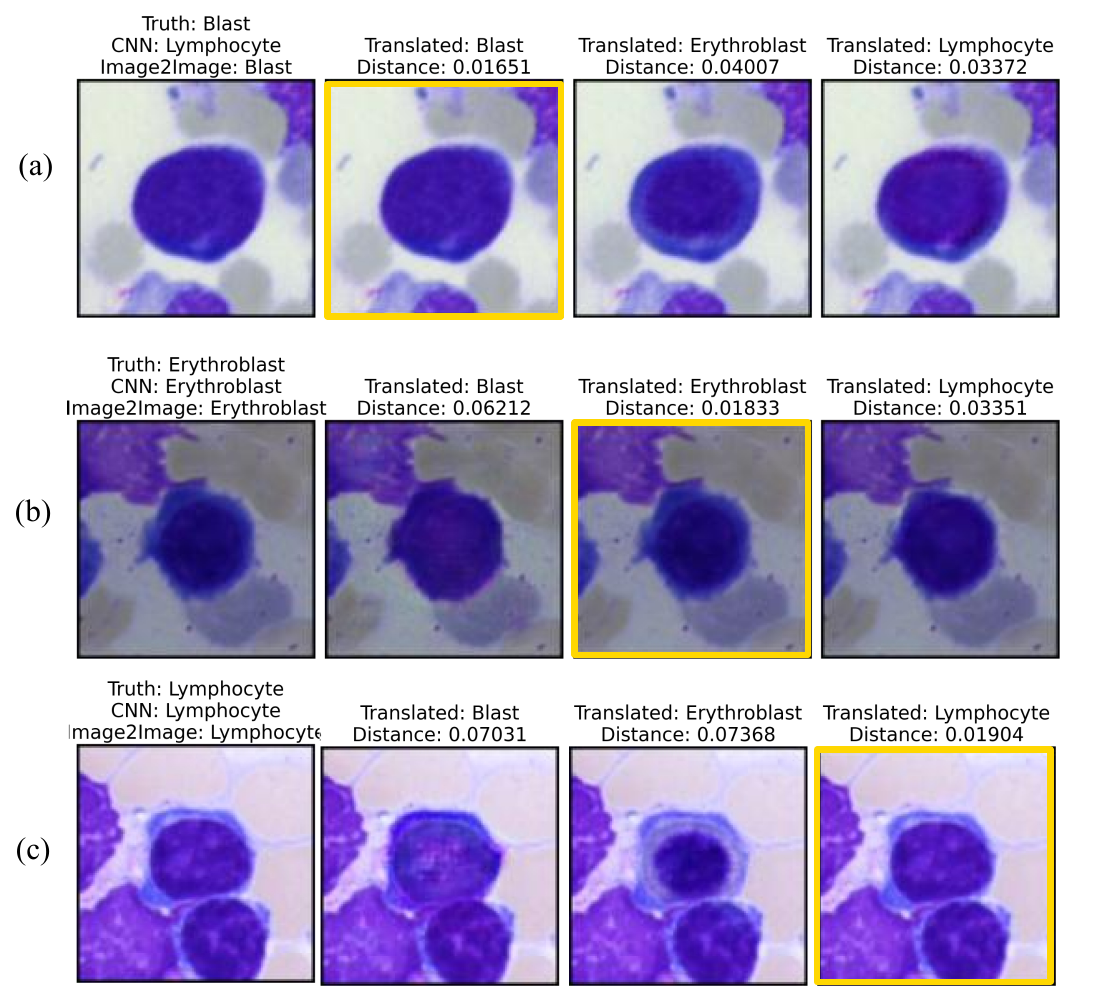}
\end{center}
   \caption{StarGAN: Bone Cytology Correct Predictions. The input image is on the far left. The $L_1$ distances are listed for each class transformation as well as the predicted class for the CNN and the translation distance SVM. Yellow highlight indicates in-class translation.}
\label{fig:samples_bone_corr}
\end{figure}

A selection of images correctly classified simply by smallest translation distance is shown in Fig.~\ref{fig:samples_bone_corr}. Note that in each of these cases, the StarGAN  primarily modified the cell at the center of the image, mostly leaving the others alone. The top row shows {a blast that was} misclassified by the baseline CNN but correctly identified by smallest translation distance. The original is darkly stained, and it is difficult to distinguish the nucleus from the cytoplasm. The blast (in-class) hypothetical retains these characteristics. The translation to an erythoblast resulted in a centered round nucleus surrounded by a lighter-stained cytoplasm, while retaining the overall cell shape. Lastly, the translation to a lymphocyte changed the overall hue to a darker purple {with a thin, lighter rim}. 

The middle row of Fig.~\ref{fig:samples_bone_corr} shows an erythroblast that was correctly {classified by both approaches}. Translation to an {erythroblast} (in-class) retains the shape of cell and nucleus, {along with} the contrast between cytoplasm and nucleus. Translation to a blast resulted in a loss distinction between the nucleus and cytoplasm, similar to the blast in the top row. Translation to a lymphocyte expanded the nucleus, {leaving a thin, lighter rim}. 

The bottom row of Fig.~\ref{fig:samples_bone_corr} shows a lymphocyte with a lighter-staining cytoplasm than the previous two examples. Translation to a lymphocyte (in-class) retains these characteristics. Translation to a blast image enlarged the nucleus and darkened the cytoplasm, consistent with the blast shown in the top row. Translation to an erythroblast, on the other hand shrank the nucleus and maintained a lighter cytoplasm. 

\begin{figure}
\begin{center}
\includegraphics[width=.8\linewidth]{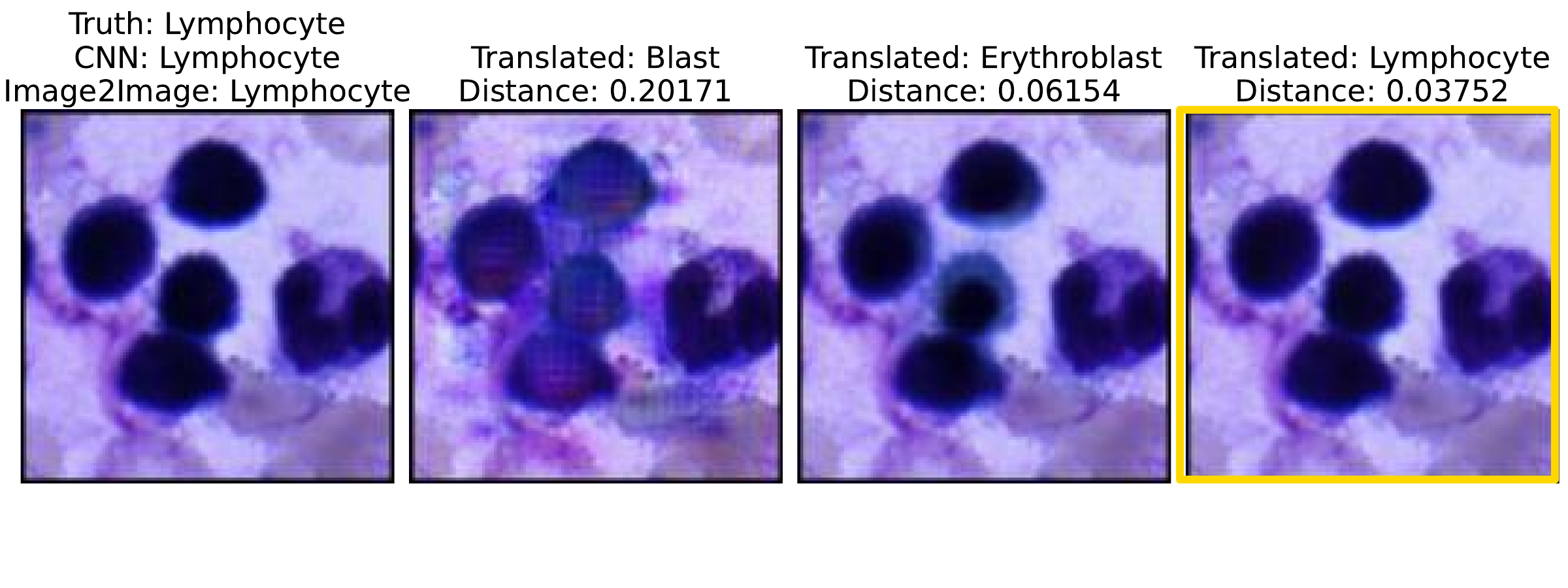}
\end{center}
   \caption{3-class StarGAN: Example bone marrow translation images showing visual artifacts. Yellow highlight indicates in-class translation.}
\label{fig:samples_bone3_wrong}
\end{figure}

 A translation failure case is shown in Fig.~\ref{fig:samples_bone3_wrong}. There are hallucinated artifacts in the produced images, especially in the translation to blast. These hallucinations alter the image in such a way that the image no longer looks realistic. However, as with the failure cases in the pigmented lesions dataset (Fig.~\ref{fig:Failure_images}), in-class translation retained the smallest distance and this error did not interfere with correct classification. 

 \subsection{Experiment 4: Bone Marrow Cytology, K=6 }
Expanding to $K=6$, the image2image network was able to achieve 90\% testing accuracy by classifying the image using the smallest translation distance, and 91.96\% with a logistic regression classifier on translation distance vectors. Individual class accuracy ranged from 86\%-97\% (Fig.~\ref{fig:logistic_reg_6_confusion}). The traditional end-to-end CNN, by comparison achieved only 86\% overall accuracy (Table~\ref{tab:translation_distance_classifiers}). We again note that compared to other traditional classifiers, ResNexT50 and EfficientNetV2\cite{HematoNet22}, the StarGAN+Translation Distance achieved improved performance for all categories except blast .

\begin{figure}
\centering
\includegraphics[width=1\linewidth] {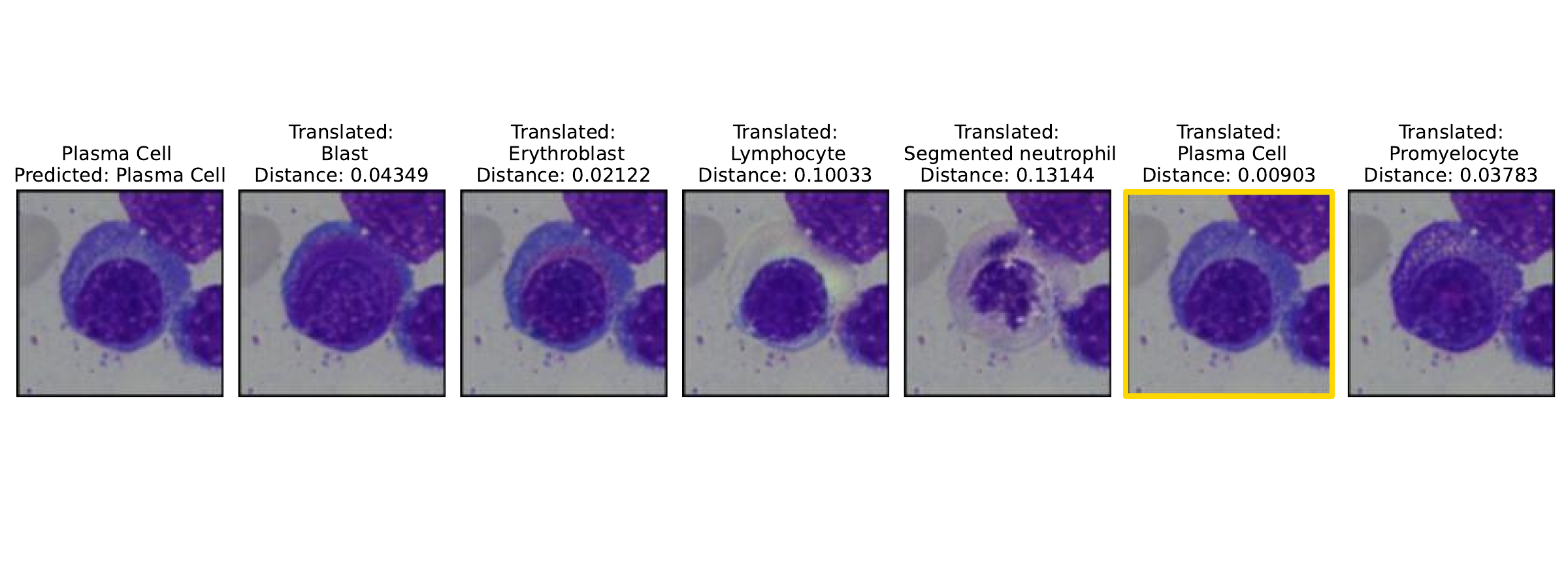}
   \caption{Experiment 4, $K=6$: Correct Image Classification. The images with the smallest translation distances are more likely to be the predicted class. The plasma cell has the smallest translation distance and thus is the predicted class. Yellow highlight indicates in-class translation.}
\label{fig:Samples6_Star}
\end{figure}

Figure~\ref{fig:Samples6_Star} shows an example image (plasma cell) correctly identified by the smallest translation distance.
In Experiment 3 ($K=3$), each cell type was morphologically similar: blasts, erythroblasts, and lymphocytes all have large, darkly stained nuclei. Lymphocytes differ by having a thin rim of lighter-stained cytoplasm. In Experiment 4 ($K=6$) case, the added cell types had more morphological variety. For instance, the samples of the segmented neutrophil tended to have a characteristic clear cell body (chromophobic cytoplasm) and segmented nuclei. As a result, the generated image for that class has more morphological changes than the other images.

\begin{figure}
\centering
\includegraphics[width=1\linewidth]{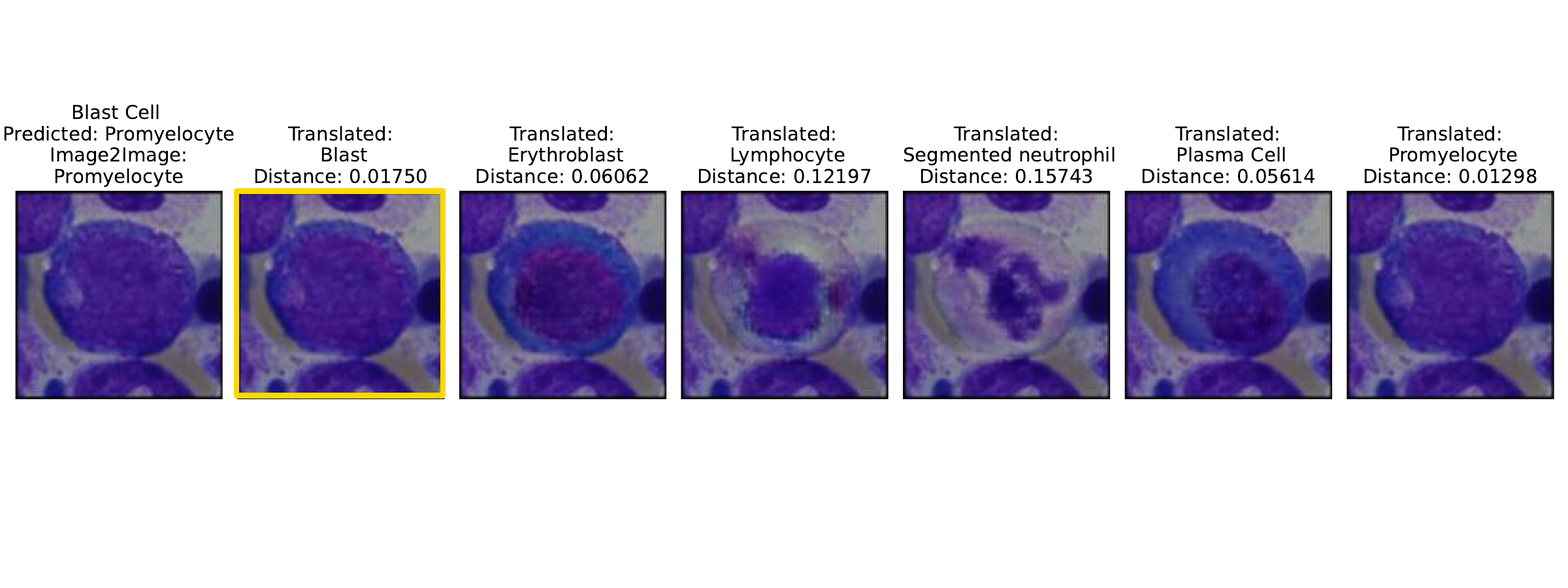}
   \caption{Experiment 4, $K=6$: Incorrect Image Classification. The predicted class is a promyleocyte due to the smallest translation distance. The next mostly likely class would be the blast. Visual inspection highlights what characteristics would need to be present for each class. Yellow highlight indicates in-class translation.}
\label{fig:Samples6_Star_wrong}
\end{figure}

{Figure~\ref{fig:Samples6_Star_wrong} shows an example image (blast) misclassified as a promelocyte by minimum translation distance. The baseline CNN misclassified this image as well, but with an alarmingly misleading confidence of 99.8\%. The advantage of the I2I-CT is readily seen here}, as a quick visual inspection shows that the hypothetical blast and promyelocyte both look visually similar to the input blast image, and both have small translation distances, indicating correctly, a much lower confidence {(roughly, a coin toss between blast and promyelocyte)} than the end-to-end CNN {asserts}.

\begin{figure}
\centering
\includegraphics[scale=0.25] {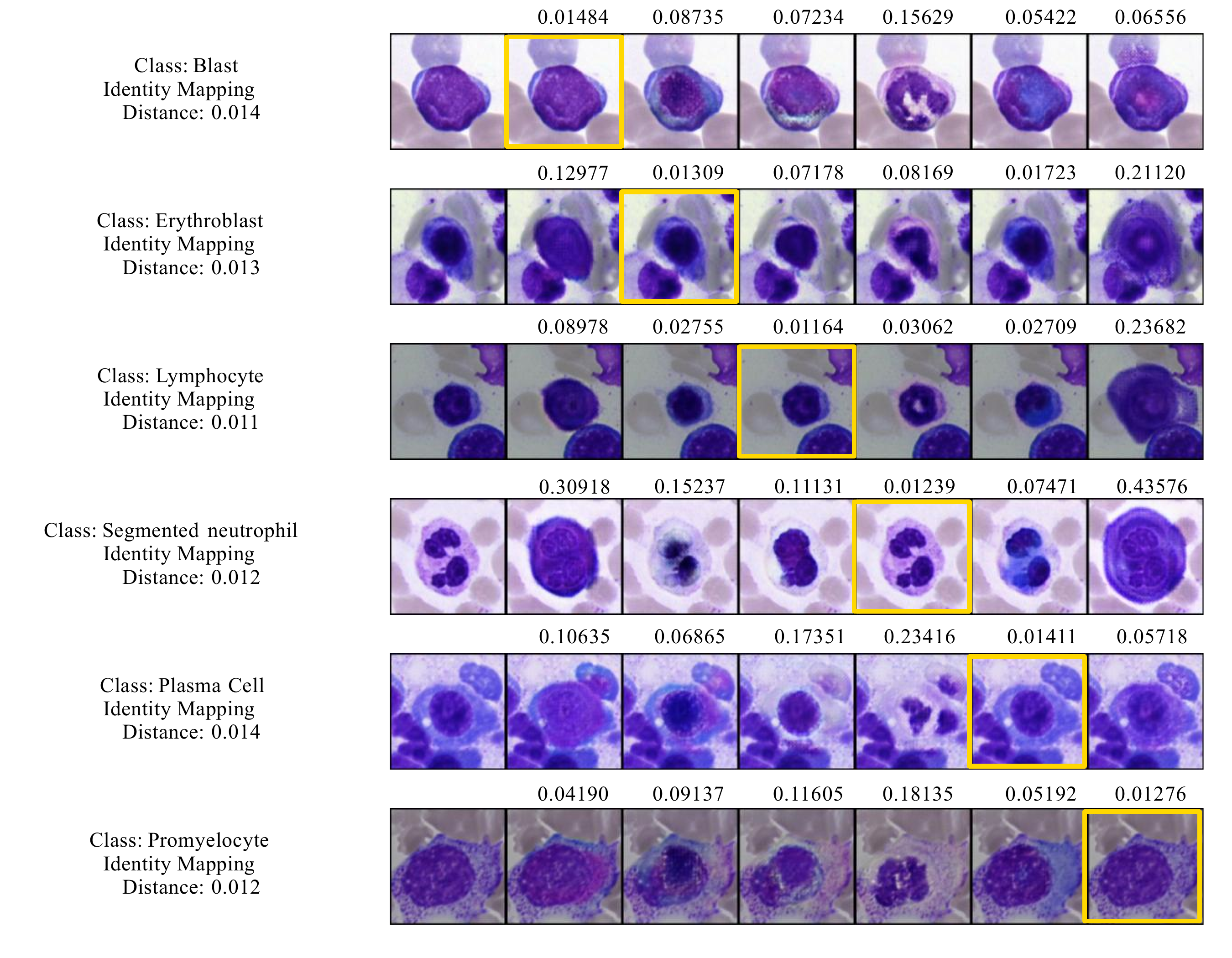}
   \caption{ Experiment 4, $K=6$ image panels. These images were correctly classified using the smallest translation distance. Yellow highlight indicates in-class translation.}
\label{fig:identity_mapping}
\end{figure}

\begin{figure}
\centering
\includegraphics[scale=0.25] {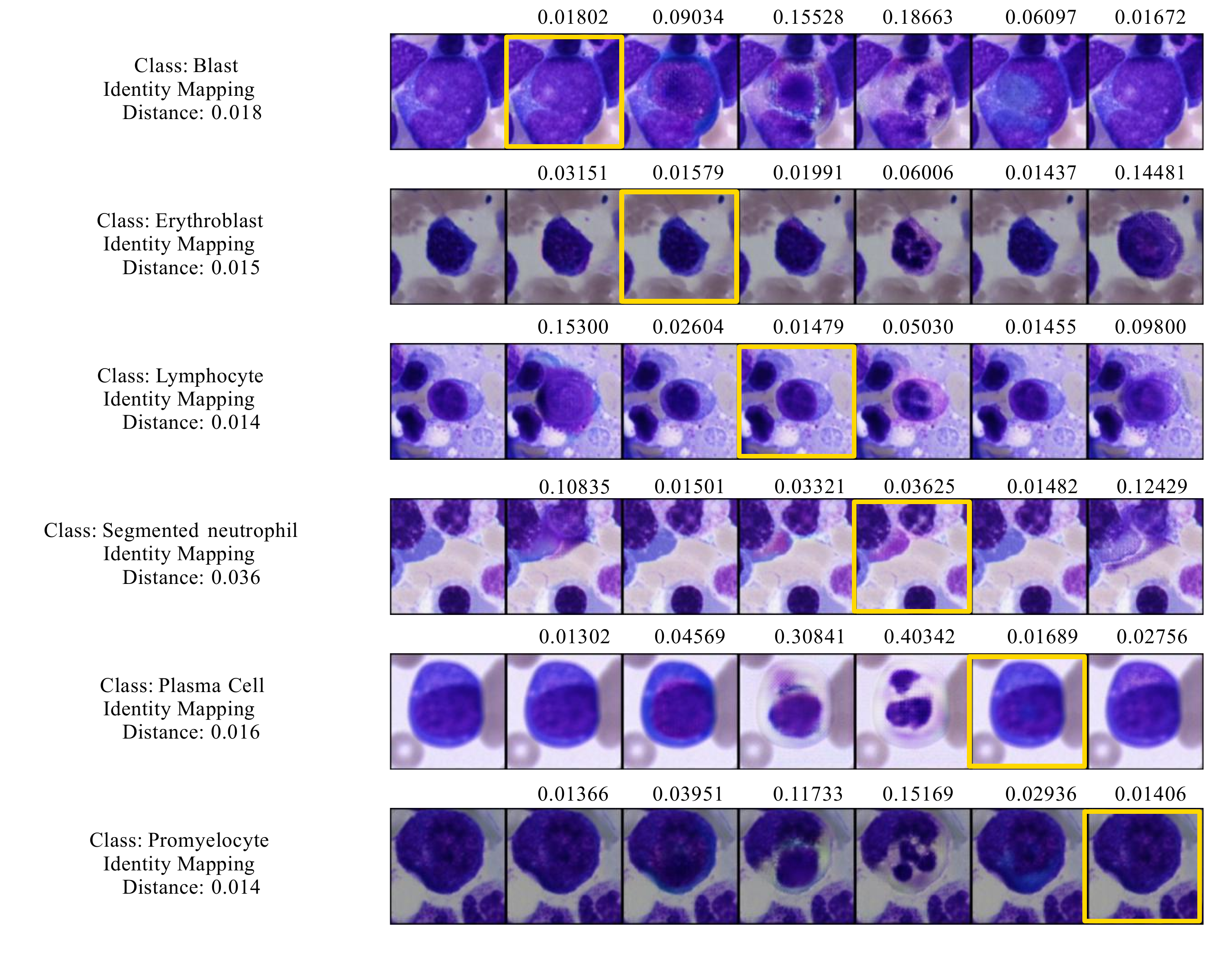}
   \caption{Experiment 4, $K=6$ image panels. These images were incorrectly classified using the smallest translation distance. Yellow highlight indicates in-class translation.}
\label{fig:identity_mapping_wrong}
\end{figure}

{Additional panels of images and I2I-CT hypotheticals are shown in Figs.~\ref{fig:identity_mapping} and ~\ref{fig:identity_mapping_wrong}. Figure~\ref{fig:identity_mapping} shows a selection of images correctly classified by smallest translation distance. As expected from the use of an identity loss, in-class translations (highlighted by yellow outlines) are visually most similar to the originals in terms of morphology, color, and texture. The most prominent morphological difference that can be seen is in the nuclei of segmented neutrophils versus other types; often, the network alters this morphology as needed, but can be seen to struggle, especially in translating the segmented neutrophil to an erythroblast and a plasma cell. By contrast, Figure~\ref{fig:identity_mapping_wrong} shows a selection of images misclassified by smallest translation distance. Typically, the in-class hypothetical (yellow outline) is still visually similar to the original, and translation distances are a close call, which indicates uncertainty. The exception is the segmented neutrophil (in-class $d_4=0.03625$, whereas $d_2=0.01501$). Yet this image appears to be an anomaly in that there is no central cell in the field of view; as noted, the network appears to have learned for the most part to alter the central cell, leaving the others untouched.}

Finally, we note that conducting the same experiment with {a random weighted sampler} to solve the class imbalance problem and achieves comparable accuracies to these experiments.

 \subsubsection{Translation Distance Classifiers }

To further improve translation distance classification, we tested a variety of simple classifiers and compared against classification by the smallest translation distance (Tab.~\ref{tab:translation_distance_classifiers}). These all had similar accuracies, with a logistic regression model with a softmax activation function performing the best (91.96\% test accuracy versus 86.07\% for the baseline end-to-end CNN). 

\begin{table}
\center
\begin{tabular}{|p{15em}|p{9em}|p{9em}|}
\hline
\textbf{Translation Distance Classifier} & \textbf{Training Accuracy (\%)} & \textbf{Testing Accuracy (\%)} \\ \hline
{Logistic Regression} & 91.92 & 91.96 \\ \hline
Scikit Logistic Regression & 91.87 & 91.90 \\ \hline
Scikit SVM Linear Kernel & 91.40 & 91.71 \\ \hline
Scikit SVM RBF Kernel & 91.42 & 91.65 \\ \hline
Scikit SVM Poly Kernel & 90.52 & 91.29 \\ \hline
{Smallest} Translation Distance $L_1$ & 91.03 & 90.07 \\ \hline
Multilayer Perceptron & 88.37 & 88.72 \\ \hline
EfficientNetB3 end-to-end & \blue{85.33} & 86.07 \\ \hline
\end{tabular}
\caption{Experiment 4, $K=6$: Accuracy versus choice of translation distance classifier for different models during training and testing. Last row shows EfficientNet results for comparison.}
\label{tab:translation_distance_classifiers}
\end{table}

\begin{figure}
\centering
\includegraphics[scale=0.55] {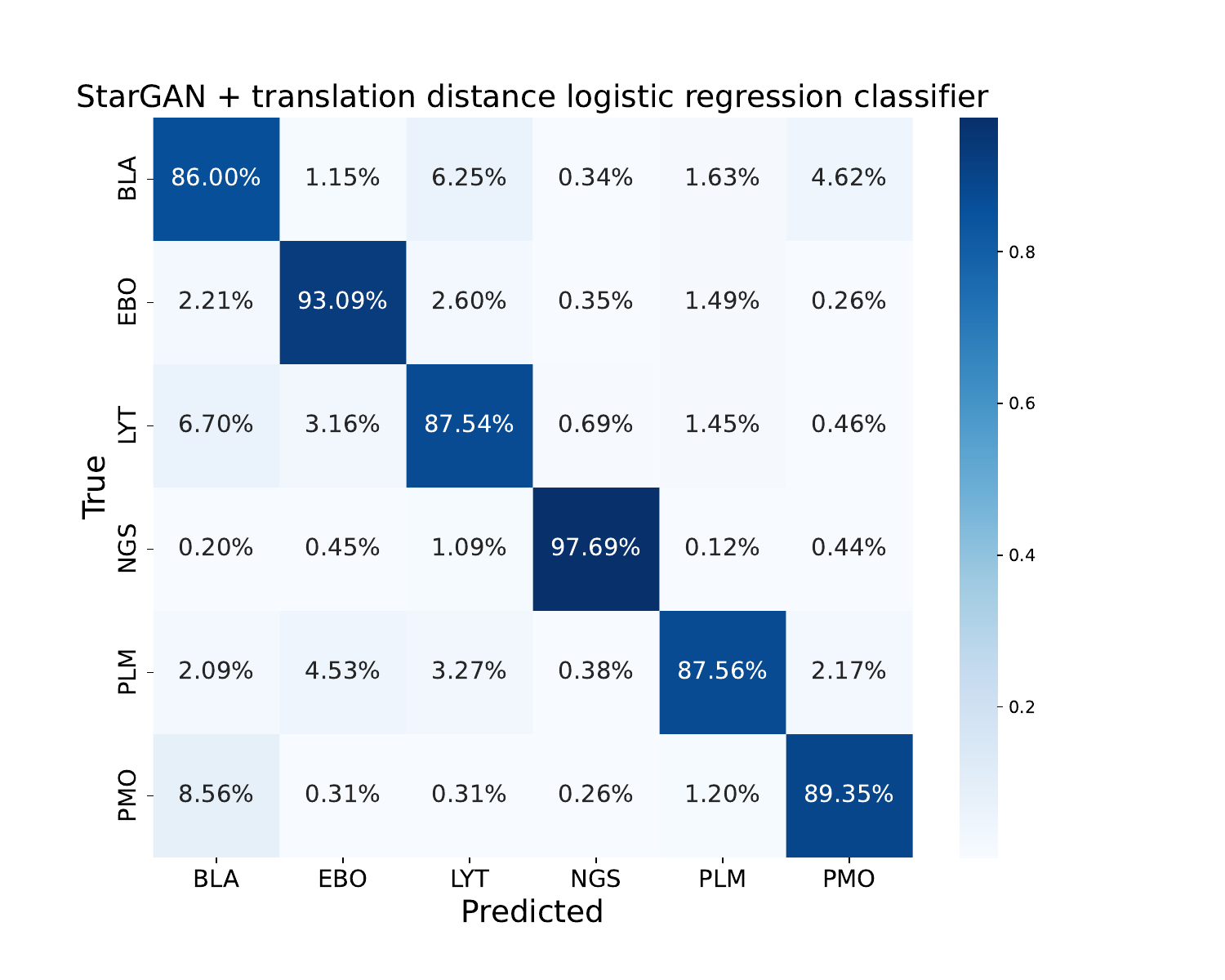}
   \caption{Experiment 4, $K=6$ confusion matrix: Logistic Regression Classifier. Higher class accuracy is represented along the diagonal.}
\label{fig:logistic_reg_6_confusion}
\end{figure}

\begin{figure}
\centering
\includegraphics[scale=0.55] {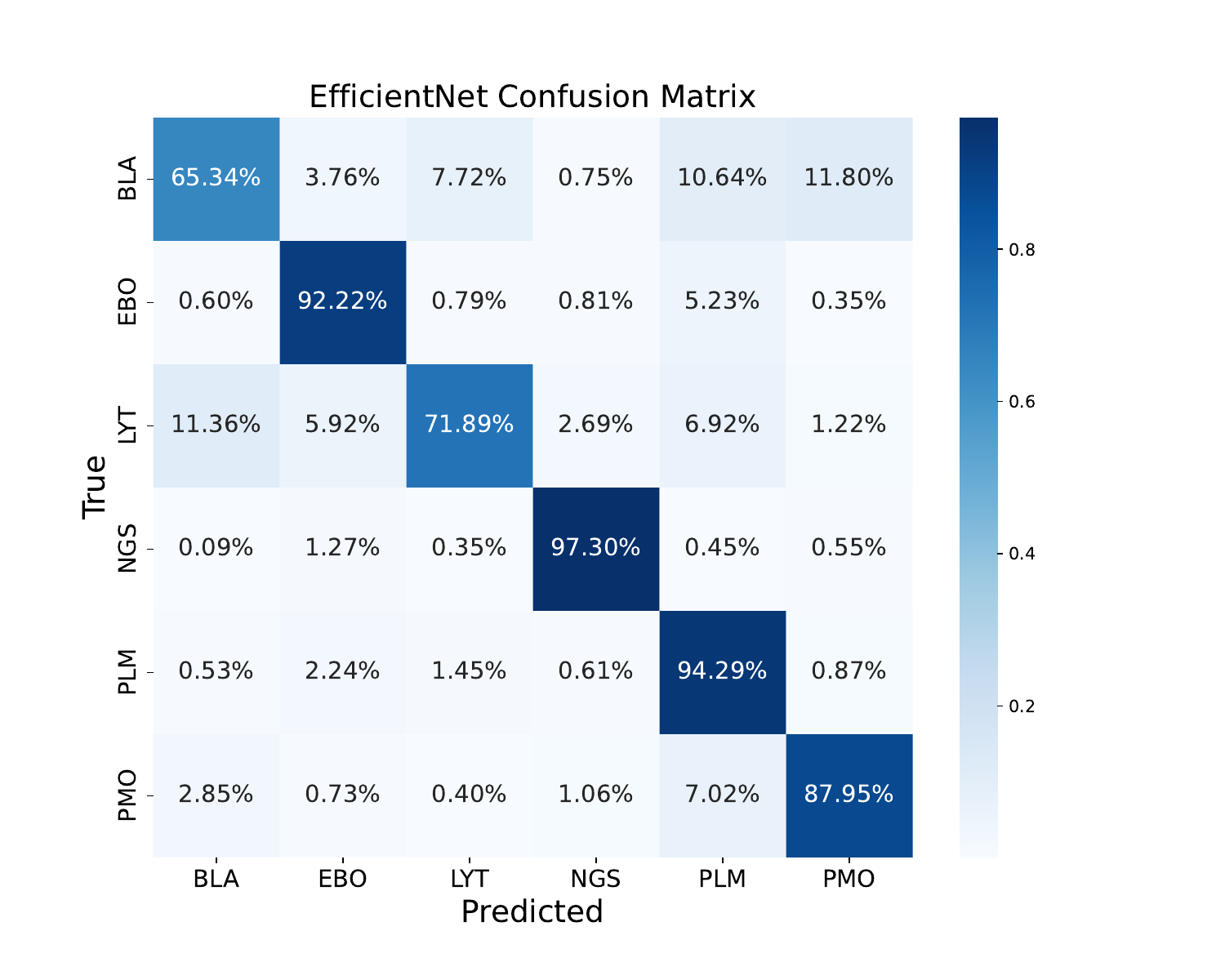}
   \caption{Experiment 4, $K=6$ confusion matrix: EfficientNetB3. Higher class accuracy is represented along the diagonal.}
\label{fig:efficient_net_6_confusion}
\end{figure}

 Figure~\ref{fig:logistic_reg_6_confusion} shows the confusion matrix for the translation distance classifier using logistic regression for probabilities. Figure~\ref{fig:efficient_net_6_confusion} shows the confusion matrix for the baseline end-to-end CNN. The highest for both was for segmented neutrophils at 97.7\% for the translation distance classifier and 97.3\% for the end-to-end CNN. The worst class accuracy for both was for blasts at 86\% for the translation distance classifier and 65\%, markedly worse for the end-to-end CNN. 

\section{Discussion} 

Using image2image networks for classification is built on the same idea behind the identity loss in CycleGANs~\cite{Zhu17}: an image translated into its own domain/class should not require any changes. Conversely, images translated into domains/classes to which they do not belong require more changes. Thus by comparing visually the source and generated images, or quantifying image translation distances it is possible to determine the class of a source image $x$. This method reduces each input image into a point in translation distance space, {$x\rightarrow d$}, which can be analyzed for clusters and trends. In this proof-of-concept work, we found $L_1$ (mean average difference) to suffice, in spite of its well-known limitations compared to other methods that better quantify perceptual differences~\cite{Brunet2012_SSIM, Zhang_2018_CVPR}. This work used a CycleGAN for $K=2$ scenarios and a StarGAN for multi-class scenarios, but we note that other networks and training approaches can in principle be used, and may lead to better results, e.g. contrastive unpaired translation \cite{park2020contrastivelearningunpairedimagetoimage}, DRIT \cite{DRIT}, cyclic autoencoders \cite{huang2020cycleconsistentadversarialautoencodersunsupervised}, transformers \cite{zheng2022ittrunpairedimagetoimagetranslation}, and stable diffusion models \cite{Rombach_2022_CVPR}.

{As with prior work in counterfactuals~\cite{DeGrave}, especially on the melanoma images~\cite{DeGrave}, we found visual inspection revealed the CNN to be inadvertently sensitive to features that a domain expert would ignore (e.g. scalebars, background pigmentation, vignetting). The predominance of such features, or of generated hallucination artifacts may be helpful in flagging potentially misclassified images-- see Figs.~\ref{fig:CNN_images},~\ref{fig:SVM_missed_images}, and~\ref{fig:Failure_images}. As with prior work in ANT-GAN on MRI images~\cite{8950113}, we were able to train a CycleGAN to transform between benign and malignant pigmented lesions. Then, instead of subtracting the source image from the generated image to highlight pixels responsible for the differences, we analyzed translation distance vectors. }

{Translation distance, unlike existing image dimension reduction methods (e.g. principal component analysis~\cite{ZhaoBaiting2021PDRM, MaJi2019Droi}, convolutional autoencoders~\cite{Encoder21}, stochastic neighbor embedding~\cite{tSNE}, which was used to compare distributions of real versus hypothetical MRI brain images in ANT~\cite{8950113}, etc.) are straightforward and interpretable. Each dimension represents the amount of change an image must undergo to fit into a given class. Ideally, the true class will show the least alteration. As we showed, the translation distance vectors have significance both for discovering trends and clusters in the data and as features in downstream classifiers.}

{On the pigmented lesions task, translation distances revealed a natural clustering along the lines of whether or not a dermatologist called for a biopsy, rather than between malignant and benign (Fig.~\ref{fig:histopathology}). This may be an interesting alternative training target for triaging pigmented lesions.} Furthermore, the overlap between benign lesions that had to be biopsied and melanomas raises the question of whether it is even possible to distinguish some types of benign lesions from melanomas. While the drive for improved accuracy often assumes all benign lesions are distinguishable from melanomas (and vice-versa) from a dermoscopy photograph alone, this is just not the case, as documented by earlier investigations in the limitations of dermoscopy~\cite{Skvara2005}. {Notably, even though pigmented lesion classifiers are known for poor generalization to external data sources~\cite{Young21,DeGrave}, the region occupied by melanomas in translation distance space (trained from ISIC 2020 data) holds up under different datasets (ISIC 2016 and 2017). Further, we went beyond prior works by showing an extension to more than two classes, and demonstrated use of the translation distance vector itself as a feature vector that can be directly used for classification.}

{Compared to a conventional end-to-end CNN classifier, we found that classification based on simple $K$-dimensional translation distance vectors achieved comparable accuracies. Two-class cases resulted in slightly worse accuracy, but 3- and 6-class cases resulted in consistently higher accuracy compared to a conventional end-to-end CNN. This could be due to differences between CycleGAN ($K=2$) and StarGAN ($K > 2$), differences between the datasets, or higher-dimensionality features of $K>2$ cases. Regardless, this is remarkable, considering, as with nearly all such end-to-end approaches, the EfficientNet's classifier operates on a feature space from the convolutional front end having $\sim 62,000$ dimensions. By comparison, $K$-dimensional translation distance vectors are a highly efficient representation apropos to classification. Further, our results from multiple years of ISIC dermoscopy data suggest translation distance features to be robust to out-of-distribution images.
In addition, our hyperparameter and architecture turning was limited to small adjustments to $\lambda_I$ and $\lambda_{\rm cycle}$. Considering the baseline EfficentNet had the advantages of extensive architecture hyperparameter tuning~\cite{Tan19} and transfer learning, its failure to significantly and consistently outperform translation distance classifiers is remarkable. This suggests the possibility that with further development, I2I-CT may enable better accuracy than currently possible. Another promising avenue is to extend the ANT-GAN approach of feeding class-specific hypotheticals along with the original image as additional input channels to a conventional classifier, which showed significant improvements in the context of MRI and brain tumor classification~\cite{8950113}. Finally, I2I-CT provided greater insights into how the network was making decisions compared to a black box classifier. In the bone cytology experiment, classification and visual errors could be more easily inspected. }

{It is also worth mentioning limitations of the I2I-CT approach. First, its complexity -- although I2I-CT produces highly compact feature vectors, it requires multiple CNNs (generator(s) and discriminator(s)) -- compared to a simple end-to-end classifier CNN, may limit applicability in cases of large images (e.g. gigapixel whole-slide images) or 3D datasets. Second, the use of CNNs for the generator(s) and discriminator(s) mean the same fundamental limitations of these architectures for end-to-end classifiers apply here. Depending on the application, though, the ability to inspect the class-specific hypotheticals and translation distance distributions may yield insights not readily apparent with end-to-end classifiers, and translation distance classifiers might in some cases yield better accuracy. }

\section{Conclusion} 
{In conclusion, image2image networks applied to translation between classes (I2I-CT) opens up unique opportunities for understanding datasets, inspecting classifier decisions, and potentially improving upon end-to-end classifiers. Generated class-specific hypothetical images can reveal training dataset artifacts that can bias or mislead conventional classifiers. The impact of such artifacts can be observed} on a case-by-case basis: an artifact's presence in the source and both generated images indicates a robust translation result, whereas an artifact's absence in one of the generated images indicates it was mistaken for a defining class-specific feature. {When multiple class-specific hypotheticals look similar to the input image, this can indicate uncertainty even in cases where an end-to-end CNN misclassifies the images with high confidence. Further, each image can be reduced to a simple length-$K$ vector of translation distances, a low-dimensional representation that can be used for visualizing image distributions, identifying natural clusters in the data, and as an input to simple classifiers. Compared to the $\sim 62,000$-dimensional feature vectors found in modern end-to-end CNN classifiers, translation distance vectors are a highly efficient representation apropos to classification. Our results on bone marrow cytology images even show the possibility of better accuracy with translation distance classifiers than end-to-end CNNs, though it remains to be seen whether this holds more broadly in other datasets.}

Finally, the image2image framework can help bridge the interpretability gap between technologies such as (AI, and microscopy techniques) with their users, clinicians. While the clinician-in-the-training-loop approach yields impressive improvements in accuracy and robustness against artifacts~\cite{DeGrave}, our approach is less labor intensive and provides clinicians an opportunity to examine visual differences between classes at inference (point of care). Both approaches are complementary and add value to AI processing and evaluation of medical imaging by working alongside, rather than to replace clinicians, and have potential for both improving diagnoses and in training human observers to recognize subtle differences between classes. {Finally, the use of image2image for classification may yield insights into difficult classification tasks and datasets that will hopefully inform efforts to produce better training datasets and in turn, improve outcomes with conventional classification CNNs.}

\section{Disclosures}
The authors declare that there are no financial interests, commercial affiliations, or other potential conflicts of interest that could have influenced the objectivity of this research or the writing of this paper.

\section {Code, and Data Availability} 
The code is available at the GitHub repository: \url{https://github.com/mikylab/image2image}, a fork of the original StarGAN repository\cite{STARGAN}. 

The Apples to Oranges data~\cite{Zhu17} is publicly available at \url{https://efrosgans.eecs.berkeley.edu/cyclegan/datasets/}. 

The ISIC Skin Lesion dataset~\cite{ISIC20} is available at \url{https://challenge2020.isic-archive.com }. 

The Bone Marrow Cytology Dataset~\cite{bone_data} is publicly available at \url{https://www.cancerimagingarchive.net/collection/bone-marrow-cytomorphology_mll_helmholtz_fraunhofer/}.

 \section{Acknowledgments}
 We thank Priya Wolff (University of Colorado School of Medicine) and Yevgeniy Seminov (Mass. General Brigham) for helpful conversations and suggestions,  {Veronica Rotemberg (Memorial Sloan Kettering Cancer Center) for discussions on lesion difficulty}, and Bill Carpenter (Colorado State University) for technical support with computational resources. Experiments were carried out on the Riviera cluster (Data Science Research Institute, Colorado State University). This work was funded by the Boettcher Foundation through a Boettcher Collaboration Grant and a Boettcher Educational Enrichment Grant.



\bibliography{report}   
\bibliographystyle{spiejour}   


\vspace{2ex}\noindent\textbf{Mikyla K. Bowen} is a 2020 Boettcher Scholar and current Master's student in computer science at Colorado State University. She joined Dr. Jesse Wilson's lab in 2020, where she focuses on applying machine learning and computer vision to imaging analysis. Prior to beginning her graduate studies, she received her Bachelor's degree in Data Science with Honors from Colorado State University.\blue{She is also the author of the fiction thriller Hidden in the Code (2026).}

\vspace{2ex}\noindent\textbf{Jesse W. Wilson} is a Boettcher Young Investigator and Associate Professor of \blue{Biomedical and Chemical Engineering}, and Electrical \& Computer Engineering at Colorado State University. Prior to joining CSU’s faculty, Jesse trained as a postdoc in Warren Warren’s lab at Duke University, and earned his PhD in Randy Bartels' lab at Colorado State University. He also serves as an Associate Editor for Biophotonics Discovery (SPIE) and Science Advances (AAAS).

\listoffigures
\listoftables

\end{spacing}
\end{document}